\documentclass[journal]{IEEEtran}
\usepackage{algorithm}
\usepackage{algpseudocode}
\usepackage{caption}
\usepackage{subcaption}
\usepackage{multirow}

\usepackage{graphicx}
\usepackage{tikz}
\usepackage{etoolbox}
\newcommand*{\lowmark}[1]{\lower.7em \hbox{\tikz\draw (0pt, 0pt)%
    circle (.5em) node {\makebox[0.5em][c]{\small #1}};}}
\robustify{\lowmark}
\usepackage{amsmath}
\usepackage{soul}
\newcommand{\uline}[1]{\ul{#1}}

\newcommand{\BOSn}[1]{{\textless S$#1$\textgreater}}
\newcommand{\EOS}{{\textless /S\textgreater}}

\ifCLASSINFOpdf
\else
\fi
\begin{document}
\title{A General Contextualized Rewriting Framework for Text Summarization}

\author{Guangsheng~Bao and Yue~Zhang,~\IEEEmembership{Member,~IEEE}%
\thanks{Manuscript submitted to TASLP.}
\thanks{Guangsheng Bao is with the School of Engineering, Westlake University, Hangzhou 310024, China (email: baoguangsheng@westlake.edu.cn).}% 
\thanks{Yue Zhang is with the School of Engineering, Westlake University, Hangzhou 310024, China, and also with the Institute of Advanced Technology, Westlake Institute of Advanced Study, Hangzhou 310024, China (email: yue.zhang@wias.org.cn).}}


\markboth{IEEE/ACM TRANSACTIONS ON AUDIO, SPEECH, AND LANGUAGE PROCESSING, VOL. XX, NO. X, XXXXX 2022}%
{Shell \MakeLowercase{\textit{et al.}}: Bare Demo of IEEEtran.cls for IEEE Journals}



\maketitle

\begin{abstract}
The rewriting method for text summarization combines extractive and abstractive approaches, improving the conciseness and readability of extractive summaries using an abstractive model. Exiting rewriting systems take each extractive sentence as the only input, which is relatively focused but can lose necessary background knowledge and discourse context. In this paper, we investigate contextualized rewriting, which consumes the entire document and considers the summary context. We formalize contextualized rewriting as a seq2seq with group-tag alignments, introducing group-tag as a solution to model the alignments,  identifying extractive sentences through content-based addressing. Results show that our approach significantly outperforms non-contextualized rewriting systems without requiring reinforcement learning, achieving strong improvements on ROUGE scores upon multiple extractors.
\end{abstract}


% \fi
\IEEEpeerreviewmaketitle

\section{Introduction}
Automatic text summarization \cite{maybury1999advances, tas2007survey} is the task that compresses input documents into shorter summaries while keeping their salient information. It has been tackled mainly by two basic methods, namely, extractive and abstractive.
The \emph{extractive} method generates a summary by extracting important text pieces (typically sentences) from a document and concatenating them to form the summary \cite{Nallapati2017, Narayan2018, Liu2019}. It has the advantage in content selection and faithfulness compared to the abstractive method \cite{Rush2015,Nallapati2016,Chopra2016}. However, the extractive sentences may contain irrelevant or redundant information \cite{Durrett2016, Chen2018rewrite, Gehrmann2019} and may have low coherence since the discourse relations and cross-sentence anaphora are not maintained \cite{Dorr2003, cheng2016neural}.
The \emph{abstractive} method uses a conditional language model to generate the summary from scratch token by token  \cite{Rush2015,Nallapati2016,See2017}, producing more fluent and readable content.
A line of work proposes a rewriting method to combine the advantage of extractive and abstractive methods. It paraphrases the extractive sentences using an abstractive model, removing irrelevant information and normalizing the expressions. Various rewriting systems have been developed, including sentence compression \cite{Durrett2016}, syntax simplification \cite{Dorr2003}, and paraphrasing \cite{Chen2018rewrite, Bae2019rewrite, Wei2019sharing, Xiao2020rewrite}. The human evaluation shows that rewriting methods can improve the readability and conciseness of extractive summaries \cite{Bae2019rewrite, Xiao2020rewrite}.

\begin{figure}[t]
    \setlength{\fboxrule}{0.5pt}
    \setlength{\fboxsep}{0.2cm}
    \fbox{\parbox{0.95\linewidth}{
        \textbf{Source Document: }
        \uline{`` Success Kid ''} is likely the Internet 's most famous baby . You 've seen him in dozens of memes , fist clenched in a determined look of persevering despite the odds . Success Kid -- now an 8-year-old named \uline{Sammy Griner} -- needs a little bit of that mojo to rub off on his family . {\it His dad , Justin , needs a kidney transplant .} About a week ago , Laney Griner , Justin 's wife and Sammy 's mother , created a GoFundMe campaign with a goal of \$ 75,000 to help cover the medical expenses that go along with a kidney transplant ...

        \vskip 0.2cm
        \textbf{Extractive Summary: }
        {\it His dad , Justin , needs a kidney transplant .}
        
        \vskip 0.2cm
        \textbf{Rewritten Summary: }
        \uline{`` Success Kid ''} \uline{Sammy Griner} 's dad , Justin , needs a kidney transplant .
        
        \vskip 0.2cm
        \textbf{Gold Summary: }
        Justin Griner , the dad of `` Success Kid , '' needs a kidney transplant .
    }}
    \caption{Key information from {\it document context} complements extractive summary, improving its informativeness.}
    \label{fig:ex-intro-a}
\end{figure}

\begin{figure*}[t]
    \setlength{\fboxrule}{0.5pt}
    \setlength{\fboxsep}{0.2cm}
    \fbox{\parbox{0.97\textwidth}{
        \setlength{\baselineskip}{15pt}    
        \textbf{Source Document: }
        ` Queen of celebrity ' is not lifetime role - before Kim Kardashian there was Paris Hilton and now there is another glossy young female fighting for the crown . \uline{A model and TV presenter known as the Mexican Kim Kardashian is hoping to dethrone the real Kim after gathering a huge support base on social media .}\lowmark{3} \uline{Jimena Sanchez , a 30 year-old Mexican model and TV presenter , is four years younger than Kim and works as a sports presenter for the Latin American division of Fox Sports , Fox Deportes .}\lowmark{1} Scroll down for video . \uline{Jimena Sanchez ( left ) is a model and TV presenter from Mexico who has been labelled the ` Mexican Kim Kardashian ' for her similarity to the reality star ( right ) .}\lowmark{2} But her real fame comes from the mass of social media attention that she gets , with one million followers on Twitter and also more than one million Facebook likes on her official page . 
        
        \vskip 0.2cm
        \textbf{Rewritten Summary: }
        \uline{Jimena Sanchez is a 30 year-old Mexican model and sports TV presenter .}\lowmark{1}
        \uline{The star is called ` Mexican Kim Kardashian ' because of their similar looks . }\lowmark{2}
        \uline{She is now hoping to become the most popular woman on social media .}\lowmark{3}
    }}
    \caption{Simple mentions to the entity in {\it summary context} simplify the complex expression of subjects in extractive sentences and improves its coherence.}
    \label{fig:ex-intro-b}
\end{figure*}

There are still issues with these rewriting methods. First, these methods generate summaries only according to the extractive sentences. Therefore, if some critical information appears in the document but not in the extractive sentences, it would be impossible for the rewriter to generate them. As an example in Fig. \ref{fig:ex-intro-a} shows, an extractive summary can be created by extracting the salient sentence   ``{\it His dad , Justin , needs a kidney transplant .}'' A rewriter can rearrange the expression but cannot resolve the entity ``{\it Sammy Griner}'' and its mention ``{\it Success Kid}'' without considering the document context during rewriting.

Second, these methods rewrite sentence by sentence independently. As a result, the cross-sentence coherence, such as entity coreference, is not modeled and controlled. Take the case in Fig. \ref{fig:ex-intro-b} as an example. The subjects of the three extractive sentences are redundant and reference to the same entity ``{\it Jimena Sanchez}'', which requires simple mentions such as ``{\it the star}'' and ``{\it she}'' in the summary sentences for conciseness and better coherence. A rewriter needs to consider the summary context of each sentence in order to remove redundancy and improve its readability.

We propose contextualized rewriting, considering both document and summary context during rewriting, rather than conditioning only on the extractive sentence. In particular, we represent extractive sentences as a part of the document representation, introducing group-tags to represent the alignment between summary and extractive sentences. Specifically, as Fig. \ref{fig:ex-intro-b} shows, we allocate the group-tags \textcircled{\scriptsize 1} \textcircled{\scriptsize 2} and \textcircled{\scriptsize 3} to the first, second, and third summary sentences. We mark the corresponding extractive sentences using the same group-tags, through which the decoder can locate the corresponding extractive sentence during rewriting. 

Based on the scheme, we propose a general framework of seq2seq with group-tag alignments for rewriting with an external or a joint internal extractor, representing sentence selection as one of the token prediction steps during decoding. This framework is independent of the implementation details of the seq2seq model, thus common for different abstractive rewriters. We instantiate three rewriters by applying the general framework to BERT \cite{devlin2018bert} and BART \cite{Lewis2019bart} models that include two rewriters with an external extractor naming BERT-Rewriter and BART-Rewriter, and one rewriter with an internal extractor naming BART-JointSR.

We evaluate the three rewriters using the popular benchmark dataset CNN/DailyMail. The results suggest that contextualized rewriting can significantly improve both the ROUGE \cite{lin2004rouge}  scores and human scores compared with previous non-contextualized rewriting methods. The performance improvement is further enlarged when contextualized rewriting works with the stronger pre-trained model BART. We further prove that the joint modeling of sentence selection and rewriting is much more beneficial than the separate pipeline model from both the efficiency and the effectiveness. To our best knowledge, we are the first rewriting method that improves ROUGE scores and human scores against both extractive and abstractive baselines.

This paper significantly extends our conference work \cite{bao2021rewrite}, which describes an instance of the general framework using BERT. The addition of this article is three-fold. First, we apply contextualized rewriting to the pre-trained model BART and demonstrate the effectiveness of our method on a stronger baseline. Second, we explore the possibility of joint modeling of sentence selection and rewriting, further powering our rewriting method by removing the dependence on an external extractor and demonstrating the advantage of joint modeling. Last, we generalize our contextualized rewriting design to a framework that suits both rewriters with an external extractor and an internal extractor, which can be applied to different seq2seq models.
Our code and models are released at https://github.com/baoguangsheng/bart-rewriters.

\begin{figure*}[t]
    \centering\small
    \includegraphics[width=0.6\linewidth]{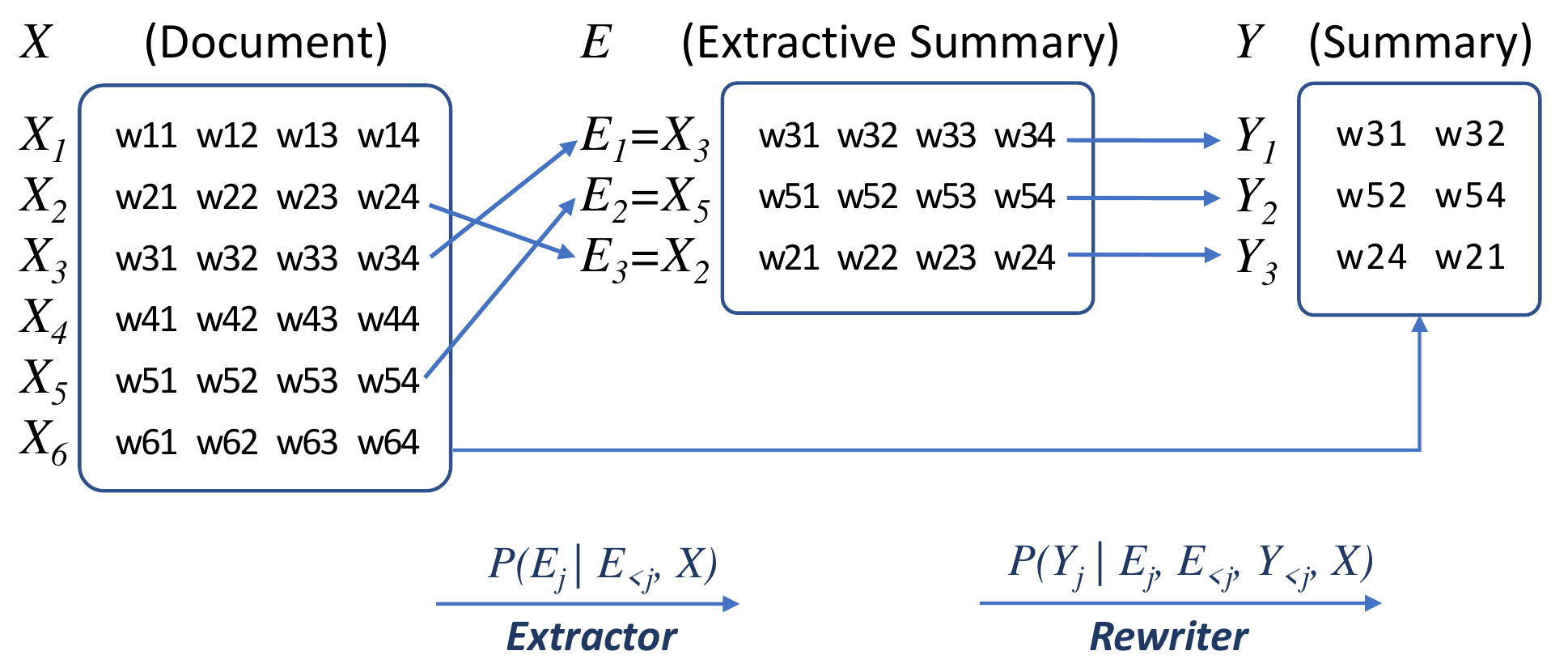}
    \caption{Contextualized rewriting involves an extractor selecting sentences from the input document and a rewriter generating a summary according to both the selected sentences and the document. (Each line in the rectangles represents a sentence, and each element w\# denotes a word.)}
    \label{fig:extractor-rewriter}
\end{figure*}

\section{Related Work}
Rewriting methods are widely used in various NLP applications, converting text from one form to another while keeping the original semantics. 
In information extraction, rewriting methods are used to generate different expressions of a query to increase the recall of retrieval \cite{dong2017learning}. 
In conversational question answering, rewriting methods are used to convert questions from a context-dependent expression to a self-contained expression with referred entities explicitly mentioned and omitted information recovered \cite{elgohary2019unpack, su2019improving, anantha2021open}. 
In neural machine translation, rewriting methods are used to rewrite the retrieved templates into translations by filling information from source sentences \cite{ren2020retrieve}.
In text summarization, rewriting methods are used to compress sentences \cite{Durrett2016learning}, simplify expressions \cite{Dorr2003hedge}, or fill templates \cite{cao2018retrieve}.
Studies \cite{Chen2018rewrite, Bae2019rewrite, Xiao2020rewrite} show that rewriting methods can enhance extractive summaries on conciseness and readability while reserving critical information.
We further develop these rewriting methods to consider document context and summary context during rewriting.

Recent studies model sentence rewriting as conditioned text generation taking extractive sentences as the inputs.
Chen and Bansal \cite{Chen2018rewrite} use an auto-regressive sentence extractor and a seq2seq model with the copy mechanism \cite{See2017} to rewrite extractive sentences one by one. They tune the extractor using reinforcement learning with reward signals from rewritten sentences and rerank rewritten summaries to avoid repetition. Bae et al. \cite{Bae2019rewrite} adopt a similar strategy but use a BERT document encoder and reward signals from the whole summary. Wei et al. \cite{Wei2019sharing} use a BERT document encoder to generate sentence embeddings and upon which a binary classifier is trained to select sentences. They use a Transformer decoder \cite{vaswani2017attention} equipped with the copy mechanism to generate summary sentences. Xiao et al. \cite{Xiao2020rewrite} use a pointer network to select sentences and make a decision of copying or rewriting accordingly. If the model chooses to rewrite, a vanilla seq2seq model will be used to rewrite the sentence. The model decisions on sentence selection and copying/rewriting are tuned using reinforcement learning. 
Compared with these methods, our rewriting method is \emph{computationally simpler} for both training and inference since we do not use reinforcement learning and the copy mechanism as above. Furthermore, as mentioned earlier, in contrast to these pure sentence rewriting methods, we consider the \emph{document context} and the \emph{summary context} during rewriting, thereby improving information recall and cross-sentence coherence. In addition, we propose a joint model of internal extractor and rewriter, reducing the total model size, the training cost, and the inference cost of the pipeline approach.

Some hybrid extractive and abstractive summarization models align with our work.
Hsu et al. \cite{hsu2018unified} propose inconsistency loss to encourage the consistency between word-level and sentence-level attentions.
Cheng and Lapata \cite{cheng2016neural} extract important words first and use them to constrain a language model to generate a summary. Gehrmann et al. \cite{Gehrmann2019} propose a bottom-up method, selecting important words using a neural classifier and restricting the copy source of a pointer-generator network to the chosen words to generate a summary. Similar to our method, they use extracted content to guide the abstractive summarizer. However, different from their methods, which focus on word-level, we investigate {\it sentence-level} constraints for guiding abstractive {\it rewriting}.

Our rewriting method can also be regarded as an abstractive summarizer \cite{Rush2015,Nallapati2016,See2017} with attention guidance using group-tags, where the group-tags are generated from the extractive summary. In comparison with the vanilla abstractive model, the advantages are three-fold. First, extractive summaries can guide the abstractive summarizer with more salient information. Second, the training difficulty of the abstractive model can be reduced when important contents are marked in inputs. Third, the summarization procedure is made more interpretable by associating a crucial source sentence with each target sentence.

\section{Problem Formulation}
We formulate contextualized rewriting as conditioned text generation that maximizes the probability of the summary given an input document. We explore two alternatives of contextualized rewriters: a rewriter with extractive summaries provided externally and a rewriter with an internal extractor. The relative advantage of the former is generality, which can be used on top of arbitrary extractive summarizers, while the advantage of the latter is independent usability. 

Formally, we use $X$ to denote an input document and $Y$ an output summary that
\begin{equation*}
  X =\{X_i\}|_{i=1}^{|X|} \text{ and }
  Y =\{Y_j\}|_{j=1}^{|Y|},
\end{equation*}
where $X_i$ denotes each sentence in $X$ and $|X|$ the number of sentences. $Y_j$ denotes each sentence in $Y$ and $|Y|$ the number of sentences in $Y$.
Here, we use a sentence-level notation instead of word-level or token-level for the simplicity of discussions because we focus on sentence-level relations between input document, extractive summary, and rewritten summary.

\subsection{Rewriter for External Extractive Summary}
Given extractive summary $E$, our contextualized rewriter rewrites it into the final summary $Y$ that
\begin{equation}
  \hat{Y} = \arg \max_{Y} P(Y|E, X),
\label{eq:problem-def1a}
\end{equation}
where $Y$ and $E$ have the same number of sentences.

As shown in Fig. \ref{fig:extractor-rewriter}, we use external extractor to select sentences from input document $X$ and obtain extractive summary $E$ that
\begin{equation}
  E=\{E_j = X_k | X_k\in X\}|_{j=1}^{|E|},
\label{eq:problem-def1a-ext}
\end{equation}
where $|E|$ denotes the number of sentences , which is equal to $|Y|$. Each sentence $E_j$ is from a document sentence $X_k$. For instance, given a document $X=\{X_1, X_2, ..., X_{6}\}$, an extractive summary could be $E=\{E_1=X_3, E_2=X_5, E_3=X_2\}$ that the first extractive sentence $E_1$ is from the third document sentence $X_3$, and so on and so forth.

The contextualized rewriter rewrites each extractive sentence $E_j$ into a summary sentence $Y_j$ given the document context and summary context. Therefore, the problem formulated in Eq \ref{eq:problem-def1a} can be further detailed as
\begin{equation}
  \hat{Y} = \arg \max_{Y} \prod_{j=1}^{|Y|} P(Y_j|\mathop{\underline{E_j}}_{\substack{\\ \text{\color{blue}rewriting}\\ \text{\color{blue}source}}}, \mathop{\underline{E_{<j}}}_{\substack{\\ \text{\color{blue}extractive}\\ \text{\color{blue}context}}}, \mathop{\underline{Y_{<j}}}_{\substack{\\ \text{\color{blue}summary}\\ \text{\color{blue}context}}}, \mathop{\underline{X}}_{\substack{\\ \text{\color{blue}document}\\ \text{\color{blue}context}}}).
\label{eq:problem-def1b}
\end{equation}

It worth noting that contextualized rewriting as expressed in Eq \ref{eq:problem-def1b} is different from previous non-contextualized rewriters \cite{Bae2019rewrite, Wei2019sharing, Xiao2020rewrite}, which do not use contextual information but directly calculate $P(Y_j|E_j)$.

\begin{figure*}[t]
    \centering\small
    \includegraphics[width=1\linewidth]{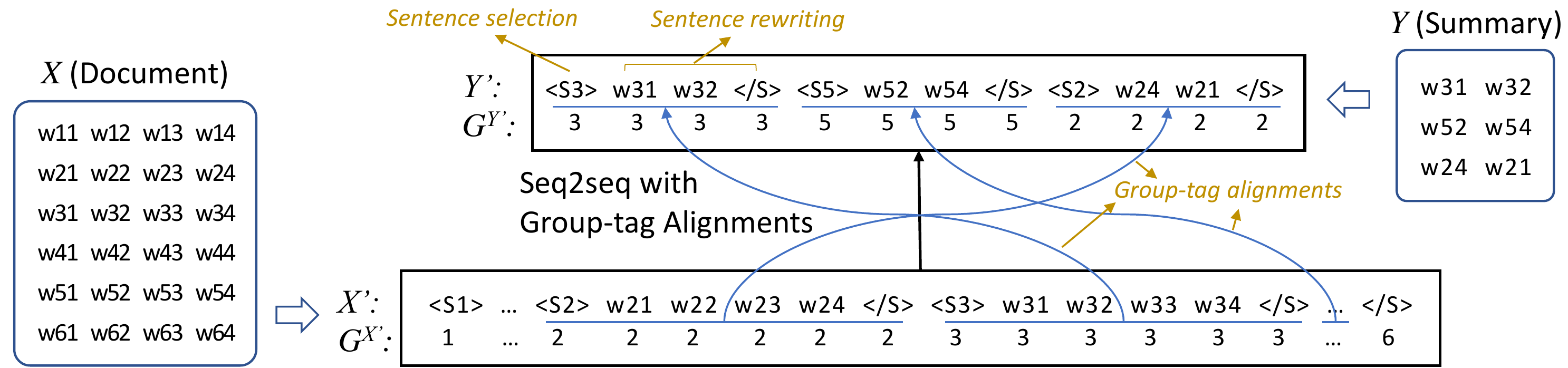}
    \caption{A general framework for contextualized rewriting which involves group-tag alignments, sentence selection, and sentence rewriting.}
    \label{fig:general-framework}
\end{figure*}

\subsection{Joint Internal Extractor and Rewriter}
The external extractor can be integrated into contextualized rewriter, resulting in a self-contained rewriter. The self-contained rewriter models both sentence extraction and rewriting in one seq2seq model. The joint problem can be defined as
\begin{equation}
  \hat{Y} = \arg_Y \max_{Y, E} P(Y, E|X),
\label{eq:problem-def2}
\end{equation}
where $E$ is the selected sentences from document $X$, which is defined as the same as Eq \ref{eq:problem-def1a-ext}. The equation can be further expanded as
\begin{equation}
  \hat{Y} = \arg_Y \max_{Y, E} \prod_{j=1}^{|Y|} P(Y_j, E_j|Y_{<j}, E_{<j}, X),
\label{eq:problem-def2a}
\end{equation}
where the sentence selection $E_j$ and rewriting $Y_j$ both depend on the history of selection and rewriting.

Given that the generation of $Y_j$ depends on the sentence selection of $E_j$, we separate the two decision steps as
\begin{equation}
  \hat{Y} = \arg_Y \max_{Y, E} \prod_{j=1}^{|Y|} P(Y_j|E_j, E_{<j}, Y_{<j}, X) P(E_j|E_{<j}, Y_{<j}, X),
\label{eq:problem-def2b}
\end{equation}
in which the current sentence selection $E_j$ depends on the previously selected sentences and rewritten sentences, while current sentence rewriting $Y_j$ depends on the current sentence selection and the history.

It worth noting that the internal extractor $P(E_j|E_{<j}, Y_{<j}, X)$ in Eq \ref{eq:problem-def2b} is different from the external extractor $P(E_j|E_{<j}, X)$ in Fig. \ref{fig:extractor-rewriter} because of the joint modeling of $Y$ and $E$. In addition, comparing detailed expressions in Eq \ref{eq:problem-def1b} and Eq \ref{eq:problem-def2b}, we could find that they share the same rewriting part $P(Y_j|E_j, E_{<j}, Y_{<j}, X)$, for which we will propose a general implementation in next section.

\section{A General Framework: \\ Seq2seq with Group-tag Alignments}
\label{sec:general-framework}
As shown in Fig. \ref{fig:general-framework}, we model contextualized rewriting as a seq2seq mapping problem with group-tag alignments. The alignments are expressed in two group-tag sequences, identifying each sentence alignment by the same group-tag in both sequences.
We extend both the input document and output summary with special sentence identifiers so that we can generate group-tag sequences accordingly. We transfer the sentence selection problem into a problem of predicting the sentence identifier tokens, which can be integrated into the seq2seq framework easily.

\subsection{Group-tag Alignments}
We introduce sentence identifier tokens such as \BOSn{1} and \BOSn{2} to represent the beginning of each sentence, extending document $X$ to $X'$ and summary $Y$ to $Y'$. 

\subsubsection{Group-tag Generation}
We generate group-tag sequences according to these identifiers. For a sentence beginning with \BOSn{k}, we assign group-tag $k$ to each token in that sentence. For example, we assign group-tag to each token of a document like ``\BOSn{3}/3 w31/3 w32/3 \EOS/3 \BOSn{5}/5 w52/5 w54/5 \EOS/5 ''. 
Formally, we use $tokens=\{w_i\}|_{i=1}^{N}$ to denote the mixed sequence of word tokens and identifier tokens. Given $tokens$, we generate the group tag sequence $tags=\{t_i\}|_{i=1}^{N}$ by Algorithm \ref{alg:grouptag}.

\begin{algorithm}[t]
\small
\hspace*{\algorithmicindent} {\bf Input}: $tokens=\{w_i\}|_{i=1}^{N}$\; \\
\hspace*{\algorithmicindent} {\bf Output}: $tags=\{t_i\}|_{i=1}^{N}$\;  
\begin{algorithmic}[1]
\Function{GroupTag}{tokens}
    \State $t \gets 0$ \Comment{default group-tag.}
    \For{$i \gets 1$ to $N$}
        \If{$w_i$ == ``\BOSn{k}''} 
            \State $t \gets k$ \Comment{new group-tag for new sentence.}
        \EndIf
        \State $t_i \gets t$ \Comment{group-tag for current sentence.}
        \If{$w_i$ == ``\EOS''}
            \State $t \gets 0$ \Comment{reset group-tag if sentence ends.}
        \EndIf
    \EndFor
    \State \textbf{return} $tags$
\EndFunction
\end{algorithmic}
\caption{Generate group-tags from a mixed sequence of word tokens and identifier tokens.}
\label{alg:grouptag}
\end{algorithm}

\subsubsection{Alignment Representation}
As Fig. \ref{fig:general-framework} shows, we use group-tag sequences $G^{X'}$ and $G^{Y'}$ to represent the extractive sentences $E$ and its one-one mapping to summary sentences $Y'$. Take the case in the figure as an example, the $G^{X'}=\{1, ..., 2, 2, 2, 2, 2, 2, 3, 3, 3, 3, 3, 3, ..., 6\}$ and $G^{Y'}=\{3, 3, 3, 3, 5, 5, 5, 5, 2, 2, 2, 2\}$, in which the tokens in $Y'$ corresponding to group-tag $3$ are generated by rewriting the tokens in $X'$ corresponding to the same group-tag $3$.

Given group-tag sequences $G^{X'}$ and $G^{Y'}$, we could remove $E$ from Eq \ref{eq:problem-def1b} and reformulate contextualized rewriting as
\begin{equation}
  P(Y_j|E_j, E_{<j}, Y_{<j}, X) = \prod_{k} P(Y'_k|G^{Y'}_{<k}, Y'_{<k}, G^{X'}, X'),
\label{eq:groupalign-def0}
\end{equation}
where $Y'_k$ denotes the $k$-th token in $Y'$, $Y'_{<k}$ the tokens before $k$, and $G^{Y'}_{<k}$ the group-tags before $k$ that $k$ ranges from the beginning to the end of sentence $Y_j$.
Note that the probability on the left of the equation is expressed on sentences, while that on the right is expressed on tokens.
We obtain group-tag sequence by Algorithm \ref{alg:grouptag}, that
\begin{equation*}
  G^{X'} = \textsc{GroupTag}(X'),
\end{equation*}
\begin{equation*}
  G^{Y'}_{<k} = \textsc{GroupTag}(Y'_{<k}).
\end{equation*}

\subsubsection{Group-tag Representation}
In the encoder-decoder framework, we convert $G^{X'}$ and $G^{Y'}$ into vector representations through a shared embedding table, which is randomly initialized and jointly trained with the encoder and decoder. The vector representations of $G^{X'}$ and $G^{Y'}$ are used to enrich vector representations of $X'$ and $Y'$, respectively. As a result, all the tokens tagged with $k$ in both $X'$ and $Y'$ have the same vector component, through which content-based addressing can be done by the attention mechanism \cite{garg2019}. Here, the group tag serves as a mechanism to constrain the attention from $Y'$ to the corresponding part of $X'$ during decoding. Unlike approaches that modify a seq2seq model by using rules \cite{Hsu2018, Gehrmann2019}, group tag enables the modification to be flexible and trainable.

\subsection{Document Encoding}
We generate group-tags $G^{X'}$ from the input $X'$ and apply the group-tag embeddings upon the document representation to produce the final representation
\begin{equation}
\begin{split}
  G^{X'} &= \textsc{GroupTag}(X'), \\
  x &= \textsc{Encoder}(X'), \\
  x &= x + \textsc{Emb}_{tag}(G^{X'}),  \\
\end{split}
\end{equation}
where $\textsc{Encoder}$ denotes the encoder module of a seq2seq model, and $\textsc{Emb}_{tag}$ denotes the embedding table of group-tags.

\subsection{Summary Decoding}
We extend a standard Transformer decoder with group-tag alignments. We generate group-tag sequence $G^{Y'}$ from summary $Y'$ and convert these group-tags into embeddings, adding them to token embeddings and position embeddings for input representation
\begin{equation}
\begin{split}
  G^{Y'} &= \textsc{GroupTag}(Y'), \\
  y &= \textsc{Emb}_{token}(Y') + \textsc{Emb}_{pos}(Y') + \textsc{Emb}_{tag}(G^{Y'}),  \\
\end{split}
\end{equation}
where $\textsc{Emb}_{token}$ denotes the token embedding table, $\textsc{Emb}_{pos}$ the position embedding table, and $\textsc{Emb}_{tag}$ the group-tag embedding table.

We predict the generation tokens according to encoder output and decoder history. The rewriting formula in Eq \ref{eq:groupalign-def2} can be implemented by
\begin{equation}
  P(Y'_k|G^{Y'}_{<k}, Y'_{<k}, G^{X'}, X') = \textsc{Decoder}(y_{<k}, x), \\
\label{eq:rewrite}
\end{equation}
where \textsc{Decoder} denotes Transformer decoder, $y$ represents the tagged token embeddings, and $x$ the encoder outputs. 
Because of the same vector components of group-tags in $x$ and $y$, the decoder addresses the $k$-th extracted sentence in $x$ when generating the $k$-th rewritten sentence.

\subsubsection{Sentence Selection}
As shown in Fig. \ref{fig:general-framework}, on the decoder side of self-contained rewriter, we use identifier tokens to denote the corresponding extractive sentences (e.g.\BOSn{7} and \BOSn{1}). Therefore, the problem of sentence selection is transformed into a problem of predicting sentence identifier tokens, which can be modeled as one step in the decoding sequence.
Formally, the sentence selection part in Eq \ref{eq:problem-def2b} can be transformed to
\begin{equation}
  P(E_j|E_{<j}, Y_{<j}, X) = P(Y'_k|G^{Y'}_{<k}, Y'_{<k}, G^{X'}, X'),
\label{eq:groupalign-def1}
\end{equation}
where the $k$-th token of the extended sequence $Y'$ is the identifier token of $j$-th selected sentence.

\begin{figure*}[t]
    \centering\small
    \includegraphics[width=1\linewidth]{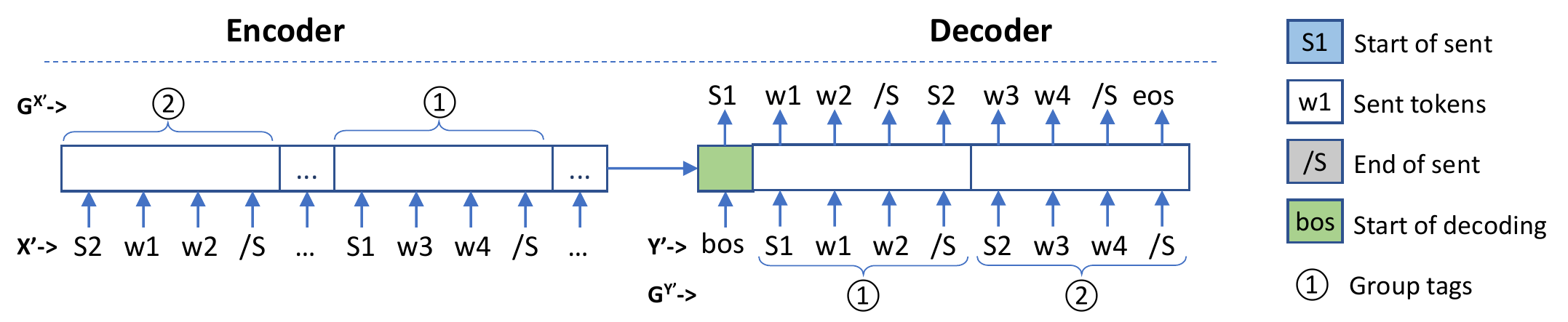}
    \caption{{\bf BERT-Rewriter} and {\bf BART-Rewriter} use extractive sentences provided by external extractor, such as the sentences denoted by the sentence identifiers \BOSn{2} and \BOSn{1} which are displayed as ``S2'' and ``S1'' in the encoder input. The sentence selection does not play a role in the models as it only predicts incremental identifiers.}
    \label{fig:bart-rewriter}
\end{figure*}

\begin{figure*}[t]
    \centering\small
    \includegraphics[width=1\linewidth]{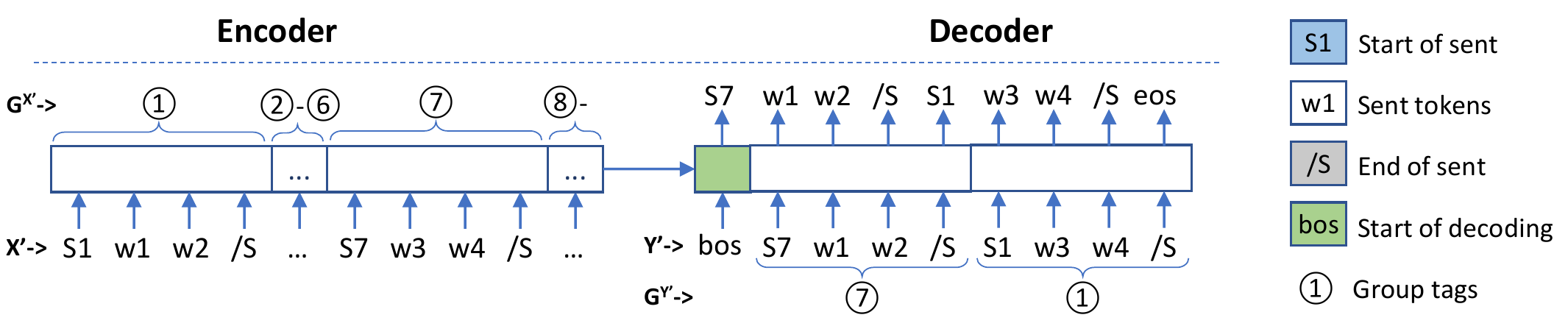}
        \caption{{\bf BART-JointSR} jointly models sentence selection and rewriting, using identifier tokens such as \BOSn{7} and \BOSn{1} (displayed as ``S7'' and ``S1'') in the decoder output to indicate the selected sentences.}
    \label{fig:bart-jointsr}
\end{figure*}

\subsubsection{Sentence Rewriting}
We can see that Eq \ref{eq:groupalign-def0} and Eq \ref{eq:groupalign-def1} have the same form and the $k$ in Eq \ref{eq:groupalign-def1} is followed by the $k$ in Eq \ref{eq:groupalign-def0}. So that we merge them into one decoding sequence, which makes sentence selection first and then do contextualized rewriting. Therefore, a general formula for both Eq \ref{eq:problem-def1b} and Eq \ref{eq:problem-def2b} is 
\begin{equation}
  \hat{Y'} = \arg \max_{Y'} \prod_{k=1}^{K} P(Y'_k|G^{Y'}_{<k}, Y'_{<k}, G^{X'}, X'),
\label{eq:groupalign-def2}
\end{equation}
where $K$ denotes the number of tokens in summary $Y'$, and $X'$ is extended from $X$ while $Y'$ is extended from $Y$.

The formula is the same for the rewriter with an external extractive summary and the rewriter with an internal extractor, but the group-tag sequences are different. For the former, the group-tag sequence $G^{X'}$ tells where the extractive sentences are located in $X'$. In contrast, the group-tag sequence $G^{Y'}$ is just naturally ordered and increases per sentence. For the latter, the group-tag sequence $G^{X'}$ is naturally ordered that the number increases per sentence that the group-tag sequence $G^{Y'}$ indicates the location of the extractive sentences in $X'$.

\subsection{Training and Inference}
We \emph{train} our contextualized rewriter on a pre-processed dataset labeled with oracle extractions.

\subsubsection{Oracle Extractions}
To generate oracle extraction, we match each sentence in the human summary to each document sentence, choosing the document sentence with the best matching score as the oracle extraction. Specifically, we use the average recall of ROUGE-1/2/L as the scoring function, which follows Wei et al.\cite{Wei2019sharing}.
Differing from existing work \cite{Liu2019}, which aims to find a {\it set} of sentences that maximizes ROUGE matching with the whole summary, we find the best match for each summary sentence. As a result, the number of extracted sentences is the same as the number of sentences in the human summary. This strategy is also adopted by Wei et al. \cite{Wei2019sharing} and Bae et al. \cite{Bae2019rewrite}.

\subsubsection{Loss Function}
We train contextualized rewriter using MLE loss, but with a different weight assigned to identifier tokens because they serve for sentence selection
\begin{equation}
\begin{split}
  Loss &= \sum_{j=k}^{|Y'|}  - w_j * \log P(Y'_k|G^{Y'}_{<k}, Y'_{<k}, G^{X'}, X'), \\
  w_j &= \gamma \text{ if } Y'_k \in \text{\it Indentifiers} \text{ else } 1,
\end{split}
\end{equation}
where {\it Identifiers} is the set of identifier tokens and $\gamma$ is a hyper-parameter determined by searching on the development set.

During \emph{inference}, we generate group tags incrementally with the tokens predicted. The group tag of the next decoding step is uniquely determined by the tokens that have been generated.

\section{Instances}

\subsection{Instance I: BERT Rewriter}
\label{sec:bert-rewriter}
We describe our conference work BERT rewriter \cite{bao2021rewrite} as a first instance of the framework, which follows the architecture design of Fig. \ref{fig:bart-rewriter}, using a pretrained BERT \cite{devlin2018bert} as the document encoder and a randomly-initialized Transformer \cite{vaswani2017attention} decoder as the summary decoder.
The rewriter extends the abstractive summarizer of \cite{Liu2019} with group-tag alignments between the encoder and the decoder. 

In order to adapt to the \emph{input} definition of BERT model, we convert the document input $X'$ before feeding it to BERT model. We replace the identifier tokens \BOSn{k} to a general sentence representation token \textsc{[cls]} and the end-of-sentence token {\EOS} to \textsc{[sep]}, so that the BERT encoder can consume. The randomly-initialized decoder shares the BERT token embedding table. The identifier tokens \BOSn{k} in summary are replaced with \textsc{[sep]}. During decoding, the first \textsc{[sep]} is translated as \BOSn{1}, the second \BOSn{2}, and so on.

Because we use a pre-trained model for the encoder and a randomly-initialized model for the decoder, we \emph{train} them using different learning-rate and warmup schedules that
\begin{equation}
\begin{split}
lr_{\textsc{Enc}} &= 0.002 \cdot min(step^{-0.5}, step \cdot warmup_{\textsc{Enc}}^{-1.5}), \\    
lr_{\textsc{Dec}} &= 0.2 \cdot min(step^{-0.5}, step \cdot warmup_{\textsc{Dec}}^{-1.5}). \\
\end{split}
\end{equation}

For \emph{inference}, we constrain the decoding sequence to a minimum length of 50, a maximum length of 200, a length penalty \cite{Wu2016} with $\alpha=0.95$, and a beam size of 5. During beam search, we block the paths on which a repeated trigram is generated, namely Trigram Blocking \cite{Paulus2018}.

\subsection{Instance II: BART Rewriter}
\label{sec:bart-rewriter}
As a second instance, we apply the general framework described in section \ref{sec:general-framework} to a pretrained BART \cite{Lewis2019bart}, extending it with the rewriting mechanism as Fig. \ref{fig:bart-rewriter} describes.

BART-Rewriter relies on an external extractor to select the salient source sentences. During training, the extractive sentences are selected by a matching algorithm that each summary sentence is matched to a source sentence. The group-tags for the summary follow a natural order for each sentence, while the group-tags for the source document may not. During inference, we can use BERT-Ext \cite{bao2021rewrite} to select sentences.

\subsection{Instance III: BART-JointSR (BART Joint Selector and Rewriter)}
As a third instance, we build a joint model of internal extractor (selector) and rewriter -  BART-JointSR. As Fig. \ref{fig:bart-jointsr} shows, similar to BART-Rewriter,  we apply the general framework to the pre-trained BART, but instead of an external extractor, we use an integrated extractor.

To train BART joint selector and rewriter, we generate different training inputs that sentence identifiers are generated using a different strategy. We use sentence identifiers with natural order for the input document but sentence identifiers corresponding to the document sentence for the summary. Therefore, the sentence identifiers in the decoder output indicate the sentence selections. 
During inference, we use beam-search to find the best sentence selection and rewriting combination.

\section{Experimental Setup}
We experiment on the standard benchmark dataset CNN/DailyMail \cite{Hermann2015}, a single-document summarization dataset comprising 312,085 online news articles with an average of 766 words per article and human written highlights with an average of 3.75 sentences per sample. We follow the standard splitting of Herman et al. \cite{Hermann2015}, containing 287,227 training samples, 13,368 validation samples, and 11,490 testing samples. We use the non-anonymous version and preprocess the dataset following the BART baseline \cite{Lewis2019bart}. 

We evaluate the quality of summaries using the automatic metric ROUGE \cite{lin2004rouge}, reporting ROUGE-1 (R-1), ROUGE-2 (R-2), and ROUGE-L (R-L). The ROUGE-1/2 represent the n-gram overlap between generated summary and gold reference, and ROUGE-L reflects the longest common sub-sequence between generated summary and gold reference.

\begin{table}[t]
    \centering
    \small
    \begin{tabular}{lccc}
        \hline
        \bf{Method} & \bf{ R-1 } & \bf{ R-2 } & \bf{ R-L} \\
        \hline
        \multicolumn{4}{c}{Extractive}  \\
        \hline
        LEAD-3 \cite{See2017} & 40.34 & 17.70 & 36.57 \\
        BERTSUMEXT \cite{Liu2019} & 43.25 & 20.24 & 39.63 \\
        \hline
        \multicolumn{4}{c}{Abstractive}  \\
        \hline
        BERTSUMABS \cite{Liu2019} & 41.72 & 19.39 & 38.76 \\
        BERTSUMEXTABS \cite{Liu2019} & 42.13 & 19.60 & 39.18 \\
        BART (large) \cite{Lewis2019bart} & 44.16 & 21.28 & 40.90 \\        
        \hline
        RNN-Ext+Abs+RL \cite{Chen2018rewrite} & 40.88 & 17.80 & 38.54 \\
        BERT-Hybrid \cite{Wei2019sharing} & 41.76 & 19.31 & 38.86 \\
        BERT-Ext+Abs+RL \cite{Bae2019rewrite} & 41.90 & 19.08 & 39.64 \\
        BERT+Copy/Rewrite+HRL \cite{Xiao2020rewrite} & 42.92 & 19.43 & 39.35 \\
        \hline
        \multicolumn{4}{c}{Our Models}  \\
        \hline
        BERT-Ext & 41.04 & 19.56 & 37.66 \\
         + BERT-Rewriter (base) & 43.52 & 20.57 & 40.56  \\
         + BART-Rewriter (base) & 43.52 & 20.76 & 40.61  \\
         + BART-Rewriter (large) & 44.26 & 21.23 & 41.34  \\
        BART-JointSR (large) & {\bf 44.72*} & {\bf 21.78*} & {\bf 41.70*} \\
        \hline
    \end{tabular}
    \caption{Main results on CNN/DailyMail. The best scores are in bold, and significantly better scores than BART baseline are marked with * ($p<0.001$, t-test). Ext and Abs denote extractive and abstractive models, respectively, and RL means reinforcement learning.}
    \label{tab:mainresult}
\end{table}

\section{Automatic Evaluation}
\label{sec:results}

\subsection{Main Results}
We evaluate our rewriters on the dataset CNN/DailyMail, reporting ROUGE-1/2/L (R-1/2/L). As shown in Table \ref{tab:mainresult}, we compare our model with previous extractive models, abstractive models, and rewriters.

Compared to previous BERT based extractive baseline BERTSUMEXT and abstractive baseline BERTSUMEXTABS, our BERT-Rewriter (base) outperforms them for an average of 0.5 and 1.2 ROUGE points, respectively. Compared with previous BERT based rewriters, our BERT-Rewriter (base) achieves the best scores, outperforming BERT+Copy/Rewrite+HRL for 1.0 points on average and especially 1.2 points on ROUGE-L, despite that our rewriter is purely abstractive without leveraging complex techniques such as copying mechanism and reinforcement learning. These results suggest the effectiveness of our rewriting method and the advantage of contextualized rewriting compared to non-contextualized rewriting.

Compared to the abstractive baseline BART, our BART-Rewriter (large) enhances the scores by 0.2 ROUGE points on average and especially 0.44 on ROUGE-L.
Our joint model BART-JointSR gives much better ROUGE scores than both BART and BART-Rewriter (large), outperforming them for an average of 0.6 and 0.4 points, respectively. The results demonstrate that our rewriting method works with large pre-trained model and benefits from joint modeling.

The architecture and size of pre-trained models are important factors influencing the performance of our rewriters. As we can see, our BART-Rewriter (base) has similar ROUGE scores as BERT-Rewriter (base). Although the BART-Rewriter is fine-tuned on the seq2seq pre-trained BART and the BERT-Rewriter is fine-tuned on a pre-trained BERT encoder and a randomly initialized decoder, the BART-Rewriter does not outperform the BERT-Rewriter because it has about $14\%$ fewer parameters. Our BART-Rewriter (large) gives much better scores than BART-Rewriter (base), improving the scores by 0.74, 0.47, and 0.73 on ROUGE-1/2/L, respectively. The results suggest that contextualized rewriting method can work with various pre-trained models, taking advantage of a larger model and further enhancing the model performance.

\begin{table}[t]
    \centering
    \begin{tabular}{@{}lcccc@{}}
        \hline
        {\bf Method} & {\bf R-1} & {\bf R-2} & {\bf R-L} & {\bf \#W} \\  
        \hline
        Oracle & 46.77 & 26.78 & 43.32 & 112  \\
        + BERT-Rewriter & 52.57 (+5.80) & 29.71 (+2.93) & 49.69 (+6.37) & 63  \\
        + BART-Rewriter & 55.01 (+8.24) & 32.07 (+5.29) & 52.08 (+8.76) & 64 \\
        \hline
        LEAD3 & 40.34 & 17.70 & 36.57 & 85  \\
        + BERT-Rewriter & 41.09 ({\bf +0.75}) & 18.19 ({\bf +0.49}) & 38.06 ({\bf +1.49}) & 55  \\
        + BART-Rewriter & 41.29 ({\bf +0.95}) & 18.49 ({\bf +0.79}) & 38.22 ({\bf +1.65}) & 53  \\
        \hline
        BERTSUMEXT & 42.50 & 19.88 & 38.91 & 80  \\
        + BERT-Rewriter & 43.31 ({\bf +0.81}) & 20.44 ({\bf +0.56}) & 40.33 ({\bf +1.42}) & 54  \\
        + BART-Rewriter & 43.35 ({\bf +0.85}) & 20.70 ({\bf +0.82}) & 40.55 ({\bf +1.64}) & 50  \\ 
        \hline
        BERT-Ext & 41.04 & 19.56 & 37.66 & 105  \\
        + BERT-Rewriter & 43.52 ({\bf +2.48}) & 20.57 ({\bf +1.01}) & 40.56 ({\bf +2.90}) & 66  \\
        + BART-Rewriter & 44.26 ({\bf +3.22}) & 21.23 ({\bf +1.67}) & 41.34 ({\bf +3.68}) & 66 \\
        \hline
    \end{tabular}
    \caption{Contextualized rewriters work with different extractors, enhancing their performance. The column \#W represents the average number of words in summary.}
    \label{tab:universality}
\end{table}

\subsection{Universality of Rewriter with External Extractive Summary}
Contextualized rewriters learn to compress, paraphrase, and abstract the extractive sentences into more concise and readable expressions. Although the rewriters are trained using oracle extractions, the learned rewriters are not limited to these oracle extractive sentences.

We evaluate the rewriters with various external extractors, including LEAD-3, BERTSUMEXT, BERT-Ext, and Oracle. As shown in Table \ref{tab:universality}, the rewriters enhance the quality of summaries generated by all four extractors. In particular, working with the basic extractor LEAD-3, BERT-Rewriter improves the average ROUGE score by 0.91, while BART-Rewriter improves the average score by 1.13. Even working with the best extractor BERTSUMEXT, BERT-Rewriter and BART-Rewriter enhance the average ROUGE score by 0.93 and 1.10, respectively. The improvements on ROUGE-L are most significant, which are more than 1.4 points for all extractors, suggesting a strong enhancement in fluency. 

It is worth noting that for BERTSUMEXT, we conduct decoding without Trigram Blocking \cite{Paulus2018}, which blocks beam search paths containing repeated trigrams. As a result, the extractive summaries contain more redundant information. However, when we apply our rewriters to them, the redundancy can be reduced and achieve higher scores.

\subsection{Advantage of Joint Internal Extractor and Rewriter}
Our joint rewriter BART-JointSR (large) gives the best ROUGE scores, outperforming BART-Rewriter for 0.46, 0.55, and 0.36 on ROUGE-1/2/L, respectively. We hypothesize that the improvement comes from two advantages of the joint model compared to the pipeline model. First, BART-JointSR models sentence selection in an auto-regressive manner, therefore, the order of selected sentences is optimized, and the redundancy between sentences is reduced. Second, BART-JointSR jointly models sentence selection and rewriting, so that not only rewriting depends on selection history but vice versa. As a result, the sentence selection and rewriting are better matched with each other.

We conduct various experiments to verify our hypothesis. As shown in Table \ref{tab:impact-joint}, we first study the impact of the order of selected sentences. We alter BART-Rewriter model to enable the decoder to decide the sentence order before rewriting, naming BART-Rewriter with reorder. Reordering extractive sentences can improve the ROUGES scores by about 0.1 on average, which is although small but confirms our first hypothesis about sentence order.

\begin{table}[t]
    \centering
    \small
    \begin{tabular}{lcccc}
        \hline
        {\bf Method} & {\bf R-1} & {\bf R-2} & {\bf R-L} & {\bf \#W} \\
        \hline
        BERT-Ext & 41.04 & 19.56 & 37.66 & 105 \\
         + BART-Rewriter & 44.26 & 21.23 & 41.34 & 66 \\
         + BART-Rewriter with reorder & 44.38 & 21.31 & 41.43 & 61 \\     
        \hline
        BART-JointSR & 44.72 & 21.78 & 41.70 & 63 \\
         - extractive summaries &  39.45 & 18.37 & 35.75 & 110 \\
         - dedup extractive summaries  & 42.43 & 20.06 & 38.64 & 93 \\
        \hline
        BART-JointSR with two-stage &  43.57 & 20.78 & 40.57 & 57 \\
         - extractive summaries & 37.12 & 16.81 & 33.24 & 131 \\    
         - dedup extractive summaries & 42.06 & 19.62 & 38.04 & 88 \\
        \hline
    \end{tabular}
    \caption{Joint modeling of sentence selection and rewriting further improves the performance. The performance will be much lower if we separate selection and rewriting, as the method BART-JointSR with two-stage shows.}
    \label{tab:impact-joint}
\end{table}

We further compare the independent extractor BERT-Ext and the inner auto-regressive extractor in BART-JointSR by concatenating the selected sentences. As the item ``extractive summaries'' under BART-JointSR shows, the ROUGE scores are much lower than that of BERT-Ext, which seems contradictive with our first hypothesis about reducing redundancy. However, when we remove the repetitive extractive sentences, we obtain much high scores, as the item ``dedup extractive summaries'' under BART-JointSR illustrates. Given our rewriter's firm information compression and redundancy reduction abilities, these duplicate extractive sentences are rewritten into different summary sentences without redundant information. It suggests that combining an auto-regressive extractor and a contextualized rewriter may be the key to its high performance.

\begin{figure}[t]
     \centering
     \includegraphics[width=0.8\linewidth]{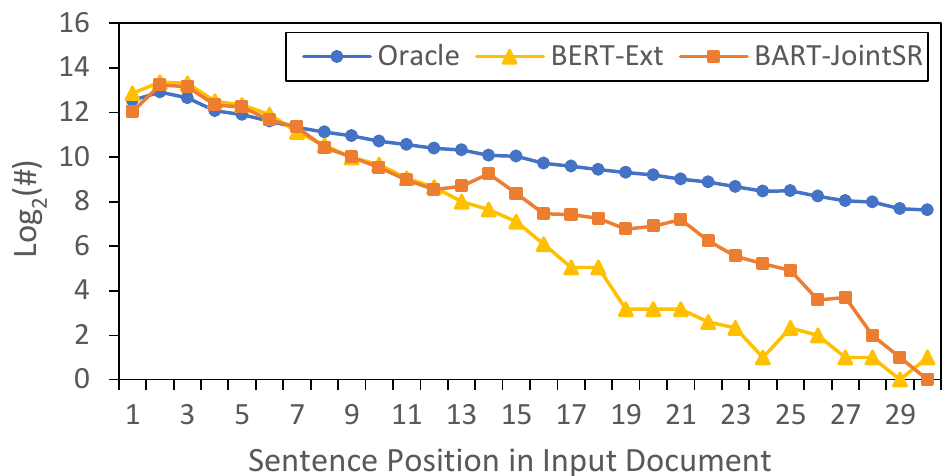}
     \caption{Distribution of extractive sentences show that the sentences selected by joint rewriter is closer to oracle extractions than BERT-Ext.}
     \label{fig:sentpos-distrib}
\end{figure}

\begin{figure}[t]
     \centering
     \includegraphics[width=0.8\linewidth]{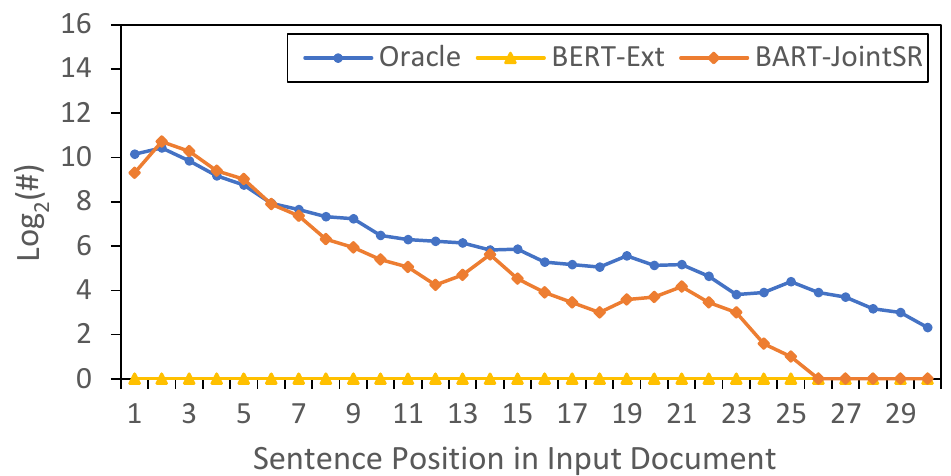}
     \caption{Distribution of {\it duplicate} extractive sentences. Joint rewriter generates duplicate extractions and the distribution is close to oracle extractions. BERT-Ext do not generate duplicate extractions, so that we use 0 to denote it.}
     \label{fig:sentpos-dup-distrib}
\end{figure}

To clarify the impact of the auto-regressive extractor, we have a closer look at the distribution of extractive sentences over the sentence position in input document. As Fig. \ref{fig:sentpos-distrib} shows, the sentences extracted by BART-JointSR have better distribution than that by BERT-Ext, which is much closer to the oracle extractions. We further observe that the oracle extractions also contain duplicate sentences, and we compare the distribution of duplicate extractive sentences in Fig. \ref{fig:sentpos-dup-distrib}. We can see that the distribution of duplicate sentences extracted by BART-JointSR is close to that of oracle extractions. In contrast, BERT-Ext only returns identical extractions because it independently makes classification decisions on each sentence.

To verify our second hypothesis about the advantage of joint modeling, we evaluate an alternative of BART-JointSR, naming BART-JointSR with two-stage, where sentence selections and sentence rewriting separated in the decoder. Specifically, for a BART-JointSR decoding sequence ``\BOSn{3} ... \BOSn{5} ... \BOSn{2} ...'', we put all sentence selection steps together at the beginning of the decoding sequence as ``\BOSn{3} \BOSn{5} \BOSn{2} {\EOS} \BOSn{3} ... \BOSn{5} ... \BOSn{2} ...''.  Therefore, sentence selection does not depend on rewriting, but rewriting still depends on sentence selection. As shown in Table \ref{tab:impact-joint}, BART-JointSR is much better than the two-stage version, outperforming it for more than 1 ROUGE point on R-1/2/L. The results suggest that the extractor can do a better job when it is conditioned on both previously selected and rewritten sentences, which strongly supports our joint modeling design.

\begin{figure*}[t]
    \centering
    \includegraphics[width=1\linewidth]{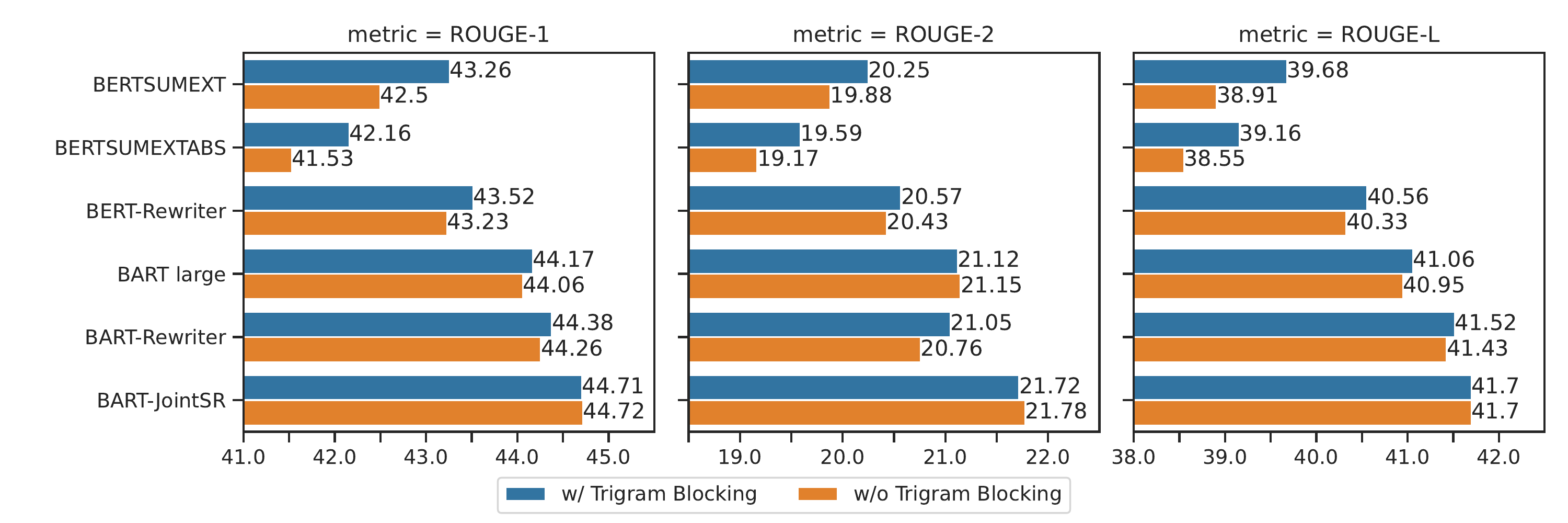}
    \caption{Models have different abilities to generate non-redundant summaries. The more redundant information is generated by the model, the bigger performance drop will happen without trigram blocking.}\label{fig:trigramblock}
\end{figure*}

\begin{table}[t]
    \centering\small
    \begin{tabular}{lcccc}
        \hline
        \bf{Method} & \bf{Info.} & \bf{Conc.} & \bf{Read.} & \bf{Faith.}  \\
        % \hline
        % RNN-Ext+Abs+RL & 3.31 & 3.23 & 3.55 & 4.63  \\
        \hline
        BERTSUMEXT & 4.01 & 3.44 & 3.41 & {\bf 5.00}  \\
        BERTSUMEXTABS & 3.87 & {\bf 3.73} & {\bf 4.06} & 4.80  \\ 
        BERT-Rewriter & {\bf 4.12} & 3.69 & 4.01 & 4.91  \\ 
        \hline
        BART & 4.19 & 3.42 & 4.27 & 4.89  \\
        BART-Rewriter & 4.31 & 3.62 & 4.21 & 4.93  \\
        BART-JointSR & {\bf 4.32} & {\bf 4.15} & {\bf 4.59} & {\bf 4.97}  \\
        \hline
    \end{tabular}
    \caption{Human evaluation on informativeness, conciseness, readability, and faithfulness.}
    \label{tab:humaneval}
\end{table}

\section{Human Evaluation}
The contextualized rewriter can improve extractive summaries in various ways. First, it can recall critical information from the document contexts, thereby increasing its informativeness. Second, it can compress unimportant/redundant information from the summaries, enhancing its conciseness. Last, it models summary discourse, therefore enhancing the readability. In addition, since the rewritten summaries are paraphrases of the extractive summaries, they tend to have fewer hallucinations than summaries generated from scratch using a pure abstractive model. We confirm these hypotheses by conducting a human evaluation. 
In particular, we randomly select 30 samples from the test set of CNN/DailyMail, scoring informativeness, conciseness, readability, and faithfulness. We assess the qualities with a number from 1(worst) to 5(best) by three independent annotators and report the average score for each quality.

The results are shown in Table \ref{tab:humaneval}. Compared to the extractive baseline BERTSUMEXT, our BERT-Rewriter improves the informativeness, conciseness, and readability by a significant margin while maintaining a high faithfulness. The improvement in informativeness mainly comes from document context,  while the improvement in conciseness and readability is mainly contributed by reduced redundancy and improved coherence. Compared with the abstractive baseline BERTSUMEXTABS and BART, our rewriter BERT-Rewriter and BART-Rewriter improve faithfulness and informativeness while keeping the conciseness and readability close.
Our joint rewriter BART-JointSR gets much higher scores than BART-Rewriter on conciseness and readability, suggesting the advantage of joint modeling of sentence selection and rewriting, which enhances the coherence between sentences.

\section{Analysis}
\label{subsec:analysis}
To better understand where the performance improvement comes from, we further conduct quantitative studies on the ability of our rewriters to reduce redundancy, avoid irrelevance, and enhance coherence.

\subsection{Redundancy}
Redundancy has been a significant problem for both extractive and abstractive summarization \cite{zhong2020matchsum, zhou2020joint}. Previous work \cite{Paulus2018, Bae2019rewrite, Liu2019sum} adopts a simple strategy during beam-search to filter out paths with duplicated n-grams. In particular, 3-gram is used by most existing methods, and the strategy is known as Trigram Blocking \cite{Paulus2018}.

As Fig. \ref{fig:trigramblock} shows, we compare two strategies for each model: one uses trigram-blocking during beam-search, and another does not. All the models except BART-JointSR give lower ROUGE scores when the trigram-blocking post-process is removed. The extractive model BERTSUMEXT depends on trigram-blocking the most; the ROUGE scores drop by 0.59 on average without it. The rewriting model BERT-Rewriter decreases the gap to 0.22 on average, suggesting that our rewriter can significantly reduce the redundant information in extractive summaries. 
Despite that trigram-blocking influences BART and BART-Rewriter for about 0.1 ROUGE point, our joint rewriter BART-JointSR gives the same scores, indicating that the joint modeling of sentence selection and rewriting can fully reduce the redundant information.

\begin{figure}[t]
    \setlength{\abovecaptionskip}{-6pt}    
    \setlength{\belowcaptionskip}{-9pt}  
     \centering
     \includegraphics[width=1\linewidth]{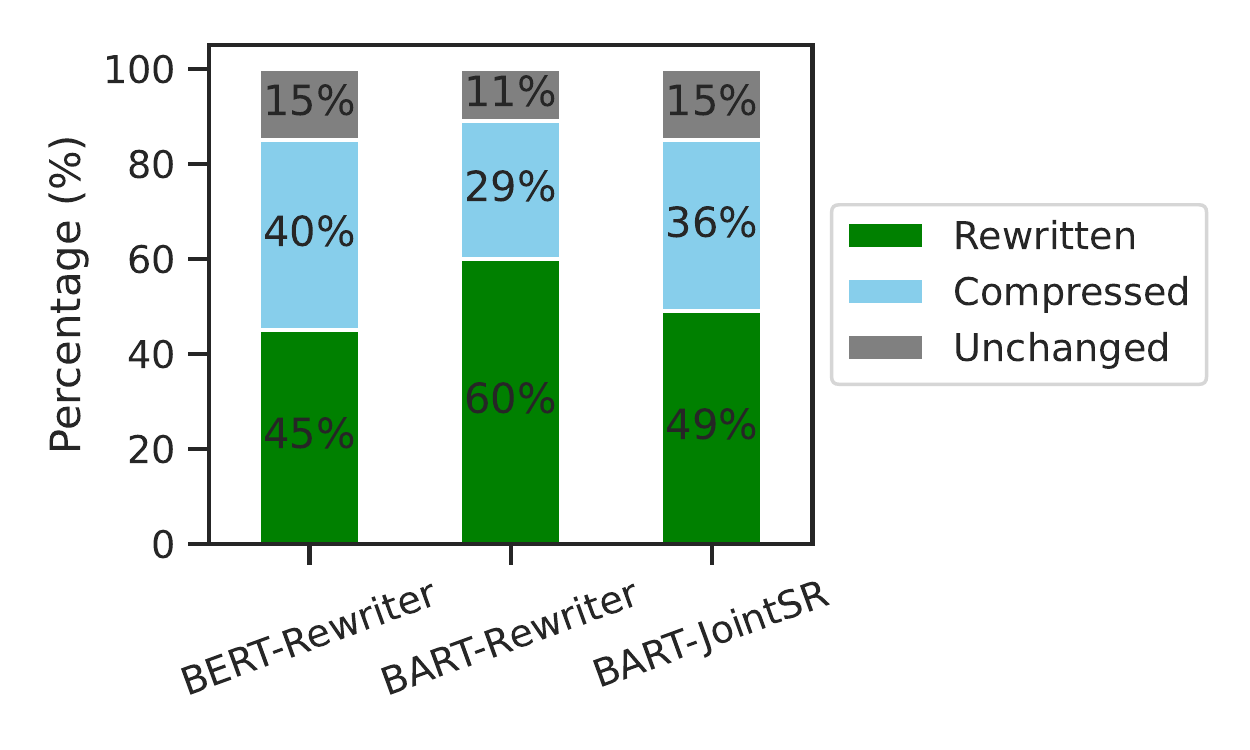}
     \caption{Proportion of rewritten, compressed, and unchanged sentences after rewriting.}\label{fig:compression}
\end{figure}

\subsection{Compression}
Contextualized rewriters can remove unimportant phrases from extractive summaries. For instance, an extractive summary ``{\it they returned to find hargreaves and the girl, who has not been named, lying on top of each other.}'' is compressed into ``{\it they returned to find hargreaves and the girl lying on top of each other.}'' According to 20 samples of BERT/BART-Rewriter and BART-JointSR generations, all the compressions are on the phrase-level instead of on single words, and most removed phrases are likely unimportant since only $11\%$ of them appear in gold summaries. 

Our contextualized rewriters can largely compress the summaries. As the column \#W in Table \ref{tab:universality} shows, the oracle summaries are compressed to almost a half by our rewriters, while other extractive summaries are compressed more than one-third.
As Fig. \ref{fig:compression} shows, most extractive sentences are rewritten or compressed. We obtain these numbers on the test dataset by adopting the edit-sequence-generation algorithm \cite{Zhang2014} to generate a sequence of word editing actions that maps an extracted summary sentence to the rewritten one. We categorize a sentence as ``Rewritten'' if the sequence contains an action of adding or modifying, ``Compressed'' if it contains an action of deleting,  and ``Unchanged'' otherwise. Take BART-JointSR as an example. Only $15\%$ of extractive sentences remain unchanged during rewriting.

% \begin{table}[t]
%     \centering\small
%     % \setlength{\abovecaptionskip}{6pt}    
%     % \setlength{\belowcaptionskip}{-9pt}    
%     \begin{tabular}{@{}lccc@{}}
%         \hline
%         \bf{Method} & \bf{1-grams} & \bf{2-grams} & \bf{3-grams} \\
%         \hline
%         GOLD & 20.66 & 56.55 & 73.48  \\
%         BERTSUMEXTABS & 1.39 & 9.81 & 17.79  \\
%         BERT-Rewriter & 1.82 & 10.74 & 19.30  \\
%         BART & 1.80 & 12.68 & 22.27  \\
%         BART-JointSR & 2.06 & 12.78 & 23.03  \\
%         BART-Rewriter & 3.07 & 17.62 & 29.86  \\
%         \hline
%     \end{tabular}
%     \caption{Percentage of novel n-grams.}
%     \label{tab:novelngrams}
% \end{table}

% \subsection{Novel n-grams}
% We measure abstractiveness using the percentage of novel n-grams. As shown in Table \ref{tab:novelngrams}, our rewriters generate summaries with higher percentage of novel n-grams than its counterpart abstractive baselines, suggesting better abstractiveness.

\begin{figure}[t]
    \setlength{\abovecaptionskip}{4pt}    
    \setlength{\belowcaptionskip}{-9pt}    
    \setlength{\fboxrule}{0.5pt}
    \setlength{\fboxsep}{0.1cm}
    \fbox{\parbox{0.95\linewidth}{
        \setlength{\baselineskip}{15pt}    
        \textbf{Source Document: }
        \uline{British world dressage champion Charlotte Dujardin won the Grand Prix at the World Cup in Las Vegas .}\lowmark{1} \uline{The 29-year-old , and her horse Valegro , who won the world title in Lyon last year , recorded a score of 85.414 percent to finish clear of Dutchman Edward Gal with American Steffen Peters in third .}\lowmark{2} Her score was short of the 87.129 she recorded in breaking her own world record last year , but there was no wiping the smile off Dujardin 's face ... The last few days , he was actually feeling not quite himself and I was a bit worried . ` But he was feeling much better and I had a really great ride . '
        
        \vskip 0.2cm
        \textbf{Rewritten Summary: }
        \uline{Charlotte Dujardin won the Grand Prix at the World Cup in Las Vegas .}\lowmark{1}
        \uline{The 29-year-old , and her horse Valegro , won the world title in Lyon last year .}\lowmark{2}

        \vskip 0.2cm
        \textbf{Swap Group Tags: }
        \uline{Charlotte Dujardin won the world title in Lyon last year .}\lowmark{1}
        \uline{The 29-year-old won the Grand Prix in Las Vegas .}\lowmark{2}
    }}
    \caption{Example of the ability to maintain coherence.}
    \label{fig:caseofcoherence}
\end{figure}

\subsection{Coherence}
Contextualized rewriters can improve extractive summaries on cohesion and coherence. It is partially illustrated by the higher ROUGE-L scores in the automatic evaluation and higher Readability scores in the human evaluation. Here, we use a case to demonstrate how our rewriter maintains the quality.

The rewriting process of our rewriters is controlled by the extractive input. By altering the extractive input and comparing the rewriting outputs, we can qualitatively study the model's ability. As Fig. \ref{fig:caseofcoherence} shows, we use different sentence orders of extractive summary to test BART-Rewriter. We can see that the athlete's name is mentioned in the first summary sentence, while a nominator is used in the second sentence. When we change the order of the two extractive sentences, as the ``Swap Group Tags'' section shows, the content of the two summary sentences interchange their positions. However, the athlete's name is still presented in the first sentence, and a nominator is used in the second sentence. These observations demonstrate that our rewriter maintains the cross-sentence anaphora correctly.

\section{Conclusion}
We investigated contextualized rewriting for automatic summarization, building both rewriters with an external extractor and an internal extractor, by proposing a general framework naming seq2seq with group-tag alignments and implement three rewriter instances. Results on standard benchmark show that contextual information is highly beneficial for summary rewriting, particularly when sentence selection and rewriting are jointly modeled. Our contextualized rewriters outperform existing non-contextualized rewriters by a significant margin, achieving strong ROUGE improvements upon multiple extractors for the first time. Our framework of seq2seq with group-tag alignments is general and can potentially be applied to other NLG tasks.

% \appendices
% \section{Training and Inference Settings}
% \begin{table}[t]
%     \centering
%     \small
%     \setlength{\abovecaptionskip}{6pt}    
%     \setlength{\belowcaptionskip}{-9pt}    
%     \begin{tabular}{lcccc}
%         \hline
%         {\bf Settings} & {\bf CNN/DailyMail} & {\bf XSum}  & {\bf WikiHow} \\
%         \hline
%         --beam & 2 & 5 & 3\\
%         --max-len-a & 0 & 0 & 0 \\
%         --max-len-b & 200 & 100 & 200 \\
%         --min-len & 20 & 10 & 10 \\
%         --lenpen & 1.0 & 1.0 & 1.0 \\
%         --no-repeat-ngram-size & 0 & 0 & 0 \\
%         \hline
%     \end{tabular}
%     \caption{Beam search settings for our BART based rewriters.}
%     \label{tab:beam-search}
% \end{table}

% \section{}

% \section*{Acknowledgment}

\ifCLASSOPTIONcaptionsoff
  \newpage
\fi



%\bibliographystyle{IEEEtran}
\bibliographystyle{IEEEtran}
\bibliography{bare_jrnl}

\begin{thebibliography}{10}
\providecommand{\url}[1]{#1}
\csname url@samestyle\endcsname
\providecommand{\newblock}{\relax}
\providecommand{\bibinfo}[2]{#2}
\providecommand{\BIBentrySTDinterwordspacing}{\spaceskip=0pt\relax}
\providecommand{\BIBentryALTinterwordstretchfactor}{4}
\providecommand{\BIBentryALTinterwordspacing}{\spaceskip=\fontdimen2\font plus
\BIBentryALTinterwordstretchfactor\fontdimen3\font minus
  \fontdimen4\font\relax}
\providecommand{\BIBforeignlanguage}[2]{{%
\expandafter\ifx\csname l@#1\endcsname\relax
\typeout{** WARNING: IEEEtran.bst: No hyphenation pattern has been}%
\typeout{** loaded for the language `#1'. Using the pattern for}%
\typeout{** the default language instead.}%
\else
\language=\csname l@#1\endcsname
\fi
#2}}
\providecommand{\BIBdecl}{\relax}
\BIBdecl

\bibitem{maybury1999advances}
M.~Maybury, \emph{Advances in automatic text summarization}.\hskip 1em plus
  0.5em minus 0.4em\relax MIT press, 1999.

\bibitem{tas2007survey}
O.~Tas and F.~Kiyani, ``A survey automatic text summarization,''
  \emph{PressAcademia Procedia}, vol.~5, no.~1, pp. 205--213, 2007.

\bibitem{Nallapati2017}
R.~Nallapati, F.~Zhai, and B.~Zhou, ``Summarunner: {A} recurrent neural network
  based sequence model for extractive summarization of documents,'' in
  \emph{Proceedings of the Thirty-First {AAAI} Conference on Artificial
  Intelligence, February 4-9, 2017, San Francisco, California, {USA}}, S.~P.
  Singh and S.~Markovitch, Eds.\hskip 1em plus 0.5em minus 0.4em\relax {AAAI}
  Press, 2017, pp. 3075--3081.

\bibitem{Narayan2018}
S.~Narayan, S.~B. Cohen, and M.~Lapata, ``Ranking sentences for extractive
  summarization with reinforcement learning,'' in \emph{Proceedings of the 2018
  Conference of the North {A}merican Chapter of the Association for
  Computational Linguistics: Human Language Technologies, Volume 1 (Long
  Papers)}.\hskip 1em plus 0.5em minus 0.4em\relax New Orleans, Louisiana:
  Association for Computational Linguistics, Jun. 2018, pp. 1747--1759.

\bibitem{Liu2019}
Y.~Liu and M.~Lapata, ``Text summarization with pretrained encoders,'' in
  \emph{Proceedings of the 2019 Conference on Empirical Methods in Natural
  Language Processing and the 9th International Joint Conference on Natural
  Language Processing (EMNLP-IJCNLP)}.\hskip 1em plus 0.5em minus 0.4em\relax
  Hong Kong, China: Association for Computational Linguistics, Nov. 2019, pp.
  3730--3740.

\bibitem{Rush2015}
A.~M. Rush, S.~Chopra, and J.~Weston, ``A neural attention model for
  abstractive sentence summarization,'' in \emph{Proceedings of the 2015
  Conference on Empirical Methods in Natural Language Processing}.\hskip 1em
  plus 0.5em minus 0.4em\relax Lisbon, Portugal: Association for Computational
  Linguistics, Sep. 2015, pp. 379--389.

\bibitem{Nallapati2016}
R.~Nallapati, B.~Zhou, C.~dos Santos, {\c{C}}.~Gul{\c{c}}ehre, and B.~Xiang,
  ``Abstractive text summarization using sequence-to-sequence {RNN}s and
  beyond,'' in \emph{Proceedings of The 20th {SIGNLL} Conference on
  Computational Natural Language Learning}.\hskip 1em plus 0.5em minus
  0.4em\relax Berlin, Germany: Association for Computational Linguistics, Aug.
  2016, pp. 280--290.

\bibitem{Chopra2016}
S.~Chopra, M.~Auli, and A.~M. Rush, ``Abstractive sentence summarization with
  attentive recurrent neural networks,'' in \emph{Proceedings of the 2016
  Conference of the North {A}merican Chapter of the Association for
  Computational Linguistics: Human Language Technologies}.\hskip 1em plus 0.5em
  minus 0.4em\relax San Diego, California: Association for Computational
  Linguistics, Jun. 2016, pp. 93--98.

\bibitem{Durrett2016}
G.~Durrett, T.~Berg-Kirkpatrick, and D.~Klein, ``Learning-based single-document
  summarization with compression and anaphoricity constraints,'' in
  \emph{Proceedings of the 54th Annual Meeting of the Association for
  Computational Linguistics (Volume 1: Long Papers)}.\hskip 1em plus 0.5em
  minus 0.4em\relax Berlin, Germany: Association for Computational Linguistics,
  Aug. 2016, pp. 1998--2008.

\bibitem{Chen2018rewrite}
Y.-C. Chen and M.~Bansal, ``Fast abstractive summarization with
  reinforce-selected sentence rewriting,'' in \emph{Proceedings of the 56th
  Annual Meeting of the Association for Computational Linguistics (Volume 1:
  Long Papers)}.\hskip 1em plus 0.5em minus 0.4em\relax Melbourne, Australia:
  Association for Computational Linguistics, Jul. 2018, pp. 675--686.

\bibitem{Gehrmann2019}
S.~Gehrmann, Y.~Deng, and A.~Rush, ``Bottom-up abstractive summarization,'' in
  \emph{Proceedings of the 2018 Conference on Empirical Methods in Natural
  Language Processing}.\hskip 1em plus 0.5em minus 0.4em\relax Brussels,
  Belgium: Association for Computational Linguistics, Oct.-Nov. 2018, pp.
  4098--4109.

\bibitem{Dorr2003}
B.~Dorr, D.~Zajic, and R.~Schwartz, ``Hedge trimmer: A parse-and-trim approach
  to headline generation,'' in \emph{Proceedings of the {HLT}-{NAACL} 03 Text
  Summarization Workshop}, 2003, pp. 1--8.

\bibitem{cheng2016neural}
J.~Cheng and M.~Lapata, ``Neural summarization by extracting sentences and
  words,'' in \emph{54th Annual Meeting of the Association for Computational
  Linguistics}.\hskip 1em plus 0.5em minus 0.4em\relax Association for
  Computational Linguistics, 2016, pp. 484--494.

\bibitem{See2017}
A.~See, P.~J. Liu, and C.~D. Manning, ``Get to the point: Summarization with
  pointer-generator networks,'' in \emph{Proceedings of the 55th Annual Meeting
  of the Association for Computational Linguistics (Volume 1: Long
  Papers)}.\hskip 1em plus 0.5em minus 0.4em\relax Vancouver, Canada:
  Association for Computational Linguistics, Jul. 2017, pp. 1073--1083.

\bibitem{Bae2019rewrite}
S.~Bae, T.~Kim, J.~Kim, and S.-g. Lee, ``Summary level training of sentence
  rewriting for abstractive summarization,'' in \emph{Proceedings of the 2nd
  Workshop on New Frontiers in Summarization}.\hskip 1em plus 0.5em minus
  0.4em\relax Hong Kong, China: Association for Computational Linguistics, Nov.
  2019, pp. 10--20.

\bibitem{Wei2019sharing}
R.~Wei, H.~Huang, and Y.~Gao, ``Sharing pre-trained {BERT} decoder for a hybrid
  summarization,'' in \emph{Chinese Computational Linguistics - 18th China
  National Conference, {CCL} 2019, Kunming, China, October 18-20, 2019,
  Proceedings}, ser. Lecture Notes in Computer Science, M.~Sun, X.~Huang,
  H.~Ji, Z.~Liu, and Y.~Liu, Eds., vol. 11856.\hskip 1em plus 0.5em minus
  0.4em\relax Springer, 2019, pp. 169--180.

\bibitem{Xiao2020rewrite}
L.~Xiao, L.~Wang, H.~He, and Y.~Jin, ``{Copy or Rewrite: Hybrid Summarization
  with Hierarchical Reinforcement Learning},'' \emph{Proceedings of the AAAI
  Conference on Artificial Intelligence (AAAI)}, 2020.

\bibitem{devlin2018bert}
J.~Devlin, M.-W. Chang, K.~Lee, and K.~Toutanova, ``Bert: Pre-training of deep
  bidirectional transformers for language understanding,'' \emph{arXiv preprint
  arXiv:1810.04805}, 2018.

\bibitem{Lewis2019bart}
M.~Lewis, Y.~Liu, N.~Goyal, M.~Ghazvininejad, A.~Mohamed, O.~Levy, V.~Stoyanov,
  and L.~Zettlemoyer, ``{BART}: Denoising sequence-to-sequence pre-training for
  natural language generation, translation, and comprehension,'' in
  \emph{Proceedings of the 58th Annual Meeting of the Association for
  Computational Linguistics}.\hskip 1em plus 0.5em minus 0.4em\relax Online:
  Association for Computational Linguistics, Jul. 2020, pp. 7871--7880.

\bibitem{lin2004rouge}
C.-Y. Lin, ``Rouge: A package for automatic evaluation of summaries,'' in
  \emph{Text summarization branches out}, 2004, pp. 74--81.

\bibitem{bao2021rewrite}
G.~Bao and Y.~Zhang, ``Contextualized rewriting for text summarization,'' in
  \emph{Proceedings of the AAAI Conference on Artificial Intelligence},
  vol.~35, no.~14, 2021, pp. 12\,544--12\,553.

\bibitem{dong2017learning}
\BIBentryALTinterwordspacing
L.~Dong, J.~Mallinson, S.~Reddy, and M.~Lapata, ``Learning to paraphrase for
  question answering,'' in \emph{Proceedings of the 2017 Conference on
  Empirical Methods in Natural Language Processing}.\hskip 1em plus 0.5em minus
  0.4em\relax Copenhagen, Denmark: Association for Computational Linguistics,
  Sep. 2017, pp. 875--886. [Online]. Available:
  \url{https://aclanthology.org/D17-1091}
\BIBentrySTDinterwordspacing

\bibitem{elgohary2019unpack}
\BIBentryALTinterwordspacing
A.~Elgohary, D.~Peskov, and J.~Boyd-Graber, ``Can you unpack that? learning to
  rewrite questions-in-context,'' in \emph{Proceedings of the 2019 Conference
  on Empirical Methods in Natural Language Processing and the 9th International
  Joint Conference on Natural Language Processing (EMNLP-IJCNLP)}.\hskip 1em
  plus 0.5em minus 0.4em\relax Hong Kong, China: Association for Computational
  Linguistics, Nov. 2019, pp. 5918--5924. [Online]. Available:
  \url{https://aclanthology.org/D19-1605}
\BIBentrySTDinterwordspacing

\bibitem{su2019improving}
\BIBentryALTinterwordspacing
H.~Su, X.~Shen, R.~Zhang, F.~Sun, P.~Hu, C.~Niu, and J.~Zhou, ``Improving
  multi-turn dialogue modelling with utterance {R}e{W}riter,'' in
  \emph{Proceedings of the 57th Annual Meeting of the Association for
  Computational Linguistics}.\hskip 1em plus 0.5em minus 0.4em\relax Florence,
  Italy: Association for Computational Linguistics, Jul. 2019, pp. 22--31.
  [Online]. Available: \url{https://aclanthology.org/P19-1003}
\BIBentrySTDinterwordspacing

\bibitem{anantha2021open}
R.~Anantha, S.~Vakulenko, Z.~Tu, S.~Longpre, S.~Pulman, and S.~Chappidi,
  ``Open-domain question answering goes conversational via question
  rewriting,'' in \emph{Proceedings of the 2021 Conference of the North
  American Chapter of the Association for Computational Linguistics: Human
  Language Technologies}, 2021, pp. 520--534.

\bibitem{ren2020retrieve}
S.~Ren, Y.~Wu, S.~Liu, M.~Zhou, and S.~Ma, ``A retrieve-and-rewrite
  initialization method for unsupervised machine translation,'' in
  \emph{Proceedings of the 58th Annual Meeting of the Association for
  Computational Linguistics}, 2020, pp. 3498--3504.

\bibitem{Durrett2016learning}
G.~Durrett, T.~Berg-Kirkpatrick, and D.~Klein, ``Learning-based single-document
  summarization with compression and anaphoricity constraints,'' in
  \emph{Proceedings of the 54th Annual Meeting of the Association for
  Computational Linguistics (Volume 1: Long Papers)}.\hskip 1em plus 0.5em
  minus 0.4em\relax Berlin, Germany: Association for Computational Linguistics,
  Aug. 2016, pp. 1998--2008.

\bibitem{Dorr2003hedge}
B.~Dorr, D.~Zajic, and R.~Schwartz, ``Hedge trimmer: A parse-and-trim approach
  to headline generation,'' in \emph{Proceedings of the {HLT}-{NAACL} 03 Text
  Summarization Workshop}, 2003, pp. 1--8.

\bibitem{cao2018retrieve}
\BIBentryALTinterwordspacing
Z.~Cao, W.~Li, S.~Li, and F.~Wei, ``Retrieve, rerank and rewrite: Soft template
  based neural summarization,'' in \emph{Proceedings of the 56th Annual Meeting
  of the Association for Computational Linguistics (Volume 1: Long
  Papers)}.\hskip 1em plus 0.5em minus 0.4em\relax Melbourne, Australia:
  Association for Computational Linguistics, Jul. 2018, pp. 152--161. [Online].
  Available: \url{https://aclanthology.org/P18-1015}
\BIBentrySTDinterwordspacing

\bibitem{vaswani2017attention}
A.~Vaswani, N.~Shazeer, N.~Parmar, J.~Uszkoreit, L.~Jones, A.~N. Gomez,
  {\L}.~Kaiser, and I.~Polosukhin, ``Attention is all you need,'' in
  \emph{Advances in neural information processing systems}, 2017, pp.
  5998--6008.

\bibitem{hsu2018unified}
W.-T. Hsu, C.-K. Lin, M.-Y. Lee, K.~Min, J.~Tang, and M.~Sun, ``A unified model
  for extractive and abstractive summarization using inconsistency loss,'' in
  \emph{Proceedings of the 56th Annual Meeting of the Association for
  Computational Linguistics (Volume 1: Long Papers)}.\hskip 1em plus 0.5em
  minus 0.4em\relax Melbourne, Australia: Association for Computational
  Linguistics, Jul. 2018, pp. 132--141.

\bibitem{garg2019}
S.~Garg, S.~Peitz, U.~Nallasamy, and M.~Paulik, ``Jointly learning to align and
  translate with transformer models,'' in \emph{Proceedings of the 2019
  Conference on Empirical Methods in Natural Language Processing and the 9th
  International Joint Conference on Natural Language Processing
  (EMNLP-IJCNLP)}.\hskip 1em plus 0.5em minus 0.4em\relax Hong Kong, China:
  Association for Computational Linguistics, Nov. 2019, pp. 4453--4462.

\bibitem{Hsu2018}
W.-T. Hsu, C.-K. Lin, M.-Y. Lee, K.~Min, J.~Tang, and M.~Sun, ``A unified model
  for extractive and abstractive summarization using inconsistency loss,'' in
  \emph{Proceedings of the 56th Annual Meeting of the Association for
  Computational Linguistics (Volume 1: Long Papers)}.\hskip 1em plus 0.5em
  minus 0.4em\relax Melbourne, Australia: Association for Computational
  Linguistics, Jul. 2018, pp. 132--141.

\bibitem{Wu2016}
Y.~Wu, M.~Schuster, Z.~Chen, Q.~V. Le, M.~Norouzi, W.~Macherey, M.~Krikun,
  Y.~Cao, Q.~Gao, K.~Macherey, J.~Klingner, A.~Shah, M.~Johnson, X.~Liu,
  {\L}.~Kaiser, S.~Gouws, Y.~Kato, T.~Kudo, H.~Kazawa, K.~Stevens, G.~Kurian,
  N.~Patil, W.~Wang, C.~Young, J.~Smith, J.~Riesa, A.~Rudnick, O.~Vinyals,
  G.~Corrado, M.~Hughes, and J.~Dean, ``{Google's Neural Machine Translation
  System: Bridging the Gap between Human and Machine Translation},'' in
  \emph{https://arxiv.org/abs/1609.08144}, 2016, pp. 1--23.

\bibitem{Paulus2018}
R.~Paulus, C.~Xiong, and R.~Socher, ``A deep reinforced model for abstractive
  summarization,'' \emph{CoRR}, vol. abs/1705.04304, 2017.

\bibitem{Hermann2015}
K.~M. Hermann, T.~Kocisk{\'{y}}, E.~Grefenstette, L.~Espeholt, W.~Kay,
  M.~Suleyman, and P.~Blunsom, ``Teaching machines to read and comprehend,'' in
  \emph{Advances in Neural Information Processing Systems 28: Annual Conference
  on Neural Information Processing Systems 2015, December 7-12, 2015, Montreal,
  Quebec, Canada}, C.~Cortes, N.~D. Lawrence, D.~D. Lee, M.~Sugiyama, and
  R.~Garnett, Eds., 2015, pp. 1693--1701.

\bibitem{zhong2020matchsum}
M.~Zhong, P.~Liu, Y.~Chen, D.~Wang, X.~Qiu, and X.-J. Huang, ``Extractive
  summarization as text matching,'' in \emph{Proceedings of the 58th Annual
  Meeting of the Association for Computational Linguistics}, 2020, pp.
  6197--6208.

\bibitem{zhou2020joint}
Q.~Zhou, N.~Yang, F.~Wei, S.~Huang, M.~Zhou, and T.~Zhao, ``A joint sentence
  scoring and selection framework for neural extractive document
  summarization,'' \emph{IEEE/ACM Transactions on Audio, Speech, and Language
  Processing}, vol.~28, pp. 671--681, 2020.

\bibitem{Liu2019sum}
Y.~Liu and M.~Lapata, ``Text summarization with pretrained encoders,'' in
  \emph{Proceedings of the 2019 Conference on Empirical Methods in Natural
  Language Processing and the 9th International Joint Conference on Natural
  Language Processing (EMNLP-IJCNLP)}.\hskip 1em plus 0.5em minus 0.4em\relax
  Hong Kong, China: Association for Computational Linguistics, Nov. 2019, pp.
  3730--3740.

\bibitem{Zhang2014}
F.~Zhang and D.~Litman, ``Sentence-level rewriting detection,'' in
  \emph{Proceedings of the Ninth Workshop on Innovative Use of {NLP} for
  Building Educational Applications}.\hskip 1em plus 0.5em minus 0.4em\relax
  Baltimore, Maryland: Association for Computational Linguistics, Jun. 2014,
  pp. 149--154.

\end{thebibliography}

% \end{thebibliography}


% \end{IEEEbiography}


%\newpage



%\vfill


\end{document}